\documentclass[final, 5p, lefttitle, sort&compress]{elsarticle}

\usepackage[usenames, dvipsnames]{color}
\usepackage{soul}
\usepackage[ruled,vlined]{algorithm2e}
\usepackage{mathtools}

\usepackage{subfigure}
\usepackage{mathtools}
\usepackage{amssymb}
\usepackage{amsfonts}
\usepackage{xcolor}
\usepackage{blindtext}

\begin{document}
\begin{frontmatter}

\title{Biomechanic Posture Stabilisation via Iterative Training of Multi-policy Deep Reinforcement Learning Agents}



\author{Mohammed~Hossny and Julie~Iskander}
\address{Deakin University\corref{lic}, Australia}
\cortext[lic]{\textcopyright 2020. This manuscript version is made available under the CC-BY-NC-ND 4.0 license http://creativecommons.org/licenses/by-nc-nd/4.0/}

\begin{abstract}
It is not until we become senior citizens do we recognise how much we took maintaining a simple standing posture for granted. It is truly fascinating to observe the magnitude of control the human brain exercises, in real time, to activate and deactivate the lower body muscles and solve a multi-link 3D inverted pendulum problem in order to maintain a stable standing posture. This realisation is even more apparent when training an artificial intelligence (AI) agent to maintain a standing posture of a digital musculoskeletal avatar due to the error propagation problem. In this work we address the error propagation problem by introducing an iterative training procedure for deep reinforcement learning which allows the agent to learn a finite set of actions and how to coordinate between them in order to achieve a stable standing posture. The proposed training approach allowed the agent to increase standing duration from 4 seconds using the traditional training method to 348 seconds using the proposed method. The proposed training method allowed the agent to generalise and accommodate perception and actuation noise for almost 108 seconds.
\end{abstract}

\begin{keyword}
\texttt{Postural Balance}\sep \texttt{Deep Reinforcement Learning}\sep \texttt{Postural Stabilisation}\sep \texttt{Fall Prevention}
\end{keyword}

\end{frontmatter}



\section{Introduction}
Human posture stabilisation is a classic control problem known as the inverted pendulum problem. While the bipedal inverted pendulum problem has been addressed with many approaches as documented in the control theory literature~\cite{Hidenori200618}, the case with the human body is slightly more complicated. This is because of the soft nature of the actuators (i.e. muscles), increased degrees of freedom (i.e. lower body joints) and its removable platform (i.e. feet). Moreover, in contrast to the common intuition, standing does require more refined levels of control than walking. This is because maintaining a standing posture requires fighting against gravity. In contrast, humans do take advantage of gravity and harness the potential energy of their upper body during walking. This is communicated by Novacheck's description of walking as controlled falling~\cite{novacheck1998biomechanics}.

\smallbreak
Modern human movement studies rely on a multidisciplinary integration between computer vision (e.g. motion capture) and biomechanics (e.g. kinesiology)~\cite{bohannon1997comfortable,serra2015postural,park2016coordination,novacheck1998biomechanics,miller1999mechanical,barroso2017combining}. While many studies rely on collecting motion capture data (MOCAP) from participants; studies investigating injuries and surgeries may not have the luxury of trying new setups on humans and thus have to rely on biomechanic simulation such as OpenSim~\cite{delp2007opensim}, Anybody~\cite{anybody2000} and, most recently, MASS~\cite{MASS2019}. These biomechanic solutions do provide a great test-bed for investigating what-if scenarios of bone deformities and muscle routing yet they have to rely on MOCAP data~\cite{delp1990interactive,delp1992force,kidzinski2020artificial}. Relying on the physics of rigid body interaction allowed these simulation environments to also simulate ocular movements as described in~\cite{iskander2018biomech,iskander2019using,iskander2018review,iskander2017sim}. 

\smallbreak
In some studies (e.g. fall detection and prevention), collecting motion data does impose ethical and safety risks on participants. In order to accommodate these challenges, researchers usually make a choice of hiring stunt actors who can fall safely or rely on 3D animation artists to generate these dangerous postures~\cite{abobakr2017fall,abobakr2018lstmfall}. Both solutions, however, do not produce realistic posture sequences and thus the need for self induced motion derived by the desire of an AI agent was highlighted in~\cite{kidzinski2018learning,Geijtenbeek,MASS2019}. In~\cite{HossnyIskander2020_dontfall}, Hossny and Iskander presented an argument for using an AI agent's failed attempts as a dataset for fall detection.



\smallbreak
In the early stages of humanoid robotics research, several control theory principals were investigated and employed. However, despite the great success of PD controllers in a wide spectrum of control problems, walking and bipedal standing remains a challenging task because of environment factors~\cite{Bottaro2005588}. This was one of the motivations deriving bio-inspired robotics. Although PD controllers remain a critical component in solving bipedal problems, the magnitude of the problem and the increased sensory information, highlighted the need for algorithms which can mimic higher levels of natural intelligence. Furthermore, the dynamic nature of the problem rendered all supervised learning approaches ineffective. This is because it is wildly impractical to collect and label enough data that can capture the different scenarios an AI agent may encounter. Thus, relying on physics based simulation engines was adopted as a viable solution. This solution was widely supported and reinforced with the recent advances in the software and hardware behind physics engines.

Over the past couple of decades, two schools of machine learning models and algorithms emerged; evolutionary algorithms (EA)~\cite{stanley2017neuroevolution} and policy gradient reinforcement learning (RL)~\cite{sutton2000policy}. 
Both approaches rely on allowing an AI agent exploration capabilities and providing a feedback score\footnote{The score is called \textit{fitness} in EA and \textit{reward} in RL} for the AI agent to optimise. The exploration component allows trying new policies to solve the problem while the obsession to optimise the feedback score allows the AI agent to fine tune the adopted policies further to actually solve the problem. In practice, evolution algorithms research has matured over the past decade and does provide viable solutions to different control problems. Perhaps the work done by Geijtenbeek et al. in~\cite{Geijtenbeek} on the locomotion of bipedal creatures using covariance matrix adaptation~\cite{HANSEN2006CMAe} is the most impressive thus far. Its major drawback, however, is purely fundamental because it is a stochastic approach and relies on selective breeding and not deriving strategies to solve the problem. This makes it a very powerful optimisation tool but lacks in providing a deterministic explanation. Reinforcement learning, on the other hand, is a deterministic approach, however it was more successful with discrete actions and lacked effective solutions for continuous control problems~\cite{mnih2013playing}. In fact, RL was considered only a powerful decision making tool until the introduction of policy gradient algorithms~\cite{lillicrap2015continuous,silver2014deterministic} which allowed deriving fine policies which can now solve continuous control problems such as the inverted pendulum problem.

\smallbreak
With the accelerating advancements in deep learning hardware and software, the policy gradient approach was further refined into the Deep Deterministic Policy Gradient (DDPG)~\cite{lillicrap2015continuous}. DDPG adopted an actor-critic architecture where both the actor and the critic are modelled as deep neural networks. The actor neural network, serves as the control module, accepts observations from the simulation/physical environment and produces actions which can affect the environment. The critic neural network is responsible for building an approximation on how the environment would behave given the current state and the action provided by the actor. Other techniques such as Soft Actor Critic (SAC)~~\cite{haarnoja2018soft,haarnoja2018softarx} and Proximal Policy Optimisation (PPO)~\cite{PPO2017} have also demonstrated promising results in continuous control problems.

\smallbreak
Typically, the actor and critic neural networks are modelled using a multi-layer-perceptron (MLP)\footnote{Depending on the environment inputs may be multidimensional where convolutional neural networks would be more suitable.}. The actor neural network accepts $N$-dimensional observation and produces $M$-dimensional action while the critic accepts $N+M$-dimensional observation/action pair and produces an estimated reward. However, for complex problems such as posture stabilisation, this typical setup suffers from one major drawback of updating the entire policy at once. This, in return, causes a rapidly changing policy every time step. Hossny and Iskander did address this issue in~\cite{HossnyIskander2020_dontfall} by introducing a modular design of the actor. In their paper, they separated observation encoding from the policies. Then they introduced the concept of using multiple policies and a coordination network to choose between them. The coordination network allows inferring a finite state machine choosing between actions produced by several policies. While the modular concept is sound, the training procedure did not take full advantage of the introduced modularity and lacked in terms of accommodating accumulating error and generalisation over environment variations. 


\smallbreak


\smallbreak
In this study we expand on Hossny and Iskander's work in~\cite{HossnyIskander2020_dontfall} to solve the generalisation problem. In doing so, we introduce an iterative training procedure. We built a similar modular actor where sub-policy neural networks are allocated and initialised randomly. We also utilised the parameterised $\tanh$ activation function to accommodate the characteristic variations of different muscles as suggested in~\cite{HossnyIskander2020_dontfall,HossnyEtal2020_PTANH}. Then, iteratively, we trained each sub-policy for a specified number of simulation steps in two stages. During the first stage of the training, both the observation sub-network and the coordinator sub-network are trainable alongside the sub-policy network. This allows the agent to constantly modify the observer and coordinator networks over all sub-policies. In the second stage of the training, we fine tune only the coordination network while locking the observer, policies, and critic networks. This forces the agent to learn how to coordinate and choose between different sub-policies according to the current state of the musculoskeletal body. We hypothesise that training on separate policies individually allows the agent to identify the actions needed for posture stabilisation as well as the coordination pattern needed to engage these sub-policies. This, in return, allows for better accommodation of propagating error and thus leads to prolonged stable standing posture for 1.8 minutes with perception and actuation noise and 5.8 minutes without noise (Supplementary Video 1).




% ...

\section{Materials and Methods}
\label{sc:methods}
\subsection{Biomechanics Simulation Environment}
The simulation environment used is based on human musculoskeletal modelled using OpenSim~\cite{kidzinski2018learning,seth2018opensim}, and OpenAI gym~\cite{brockman2016openai}, and used previously in~\cite{HossnyIskander2020_dontfall}. The musculoskeletal model~\cite{delp1990interactive,ong2017predictive,arnold2010model} is made up of seven body parts that includes the upper body, represented as a single unit, along with, two leg, each represented by three units, upper leg, lower leg and foot. The model has 14 degrees-of-freedom (DoF), 3 rotational and 3 translational DoFs for the pelvis, 2 rotational DoFs  for each hip, one rotations DoF for each knee and finally, one rotational DoF for each ankle. The model includes 22 muscles~\cite{thelen2003adjustment} for actuation, 11 muscles for each leg. The leg muscles includes the hip adductor, hip abductor, hamstrings, biceps femoris, gluteus maximus, iliopsoas, rectus femoris, vastus intermedius, gastrocnemius, soleus, and tibialis anterior. To model ground reaction forces Hunt-Crossley model is used as contact effect~\cite{hunt1975coefficient}. For each foot, contact spheres are positioned at the heel and toes and a rectangular contact plane is placed over the ground. Force is generated when the objects come into contact, which depends on the velocity of the collision and depth of penetration of the contact objects~\cite{kidzinski2018learning}.

\subsubsection{Observation Space}
The observation fed to the AI agent included 169 values covering 3D coordinates for 14 bones (42 values), 3D Euler orientation angles for 14 bones (42 values), as well as 63 values representing angles, angular velocity and angular acceleration of 7 joints (21 values each joint including the ground platform). Finally, the current muscle activation from previous state (22 values) was also included. It is important to note, that the ground reaction forces (GRFs), the centre of mass/gravity (COM/COG) and the zero moment point (ZMP) were not included in the observation space.

\subsubsection{Action Space}
The aforementioned environment was controlled by 22 continuous actions representing the muscle excitation signals of the 22 muscles. Each excitation signal has a dynamic range of 0.0 (not excited) and 1.0 (fully excited).

\subsubsection{Reward Function}
We chose a simple reward function which emphasise maintaining the pelvis height while having both feet on the ground in addition to an effort minimisation penalty. This is achieved by imposing a penalty based on the squared distance of the pelvis height ($H$) from the initial height $H_0$. Another penalty is also added for the weight distribution on both feet ($W_L, W_R$). A third penalty is added if either feet lost contact with the ground. Finally, an additional effort penalty is added to encourage the agent to reduce muscle activation ($\mathbf{m}$). The agent was rewarded each simulation time step with both feet on the ground. This is determined by the force ($\mathbf{||F_g||}$) between the ground platform and each foot separately. To that end, the reward function is formulated using the following equations.

\begin{equation}
    R_{GRF} = 
    	\begin{cases}
            1, & ||\mathbf{F_g}||>0 \\
            0, & \text{otherwise}
    	\end{cases}
    \label{eq:1}
\end{equation}

\begin{equation}
    \begin{array}{lll}
            Penalty &=& ||\mathbf{m}||^2\\
            ~&-& 16~(H-H_0)^2\\
            ~&-& 64~\left(1 - \frac{W_L}{0.65}\right)^2\\
            ~&-& 64~\left(1 - \frac{W_R}{0.65}\right)^2\\
            ~&-& 128~\left(1-R_{GRF}\right), 
        \end{array}
    \label{eq:2}
\end{equation}

\begin{equation}
    Norm.~Penalty = \frac{Penalty}{1+16+64+64+128}
    \label{eq:3}
\end{equation}

\begin{equation}
    Reward = 2~R_{GRF} - Norm.~Penalty
    \label{eq:4}
\end{equation}

Unlike the reward described in~\cite{kidzinski2020artificial} and ~\cite{HossnyIskander2020_dontfall}, we did not add a penalty for the cross leg problem. We also did not incorporate the centre of mass into the reward funnction.

\begin{figure*}
    \centering
    \includegraphics[width=.9\linewidth]{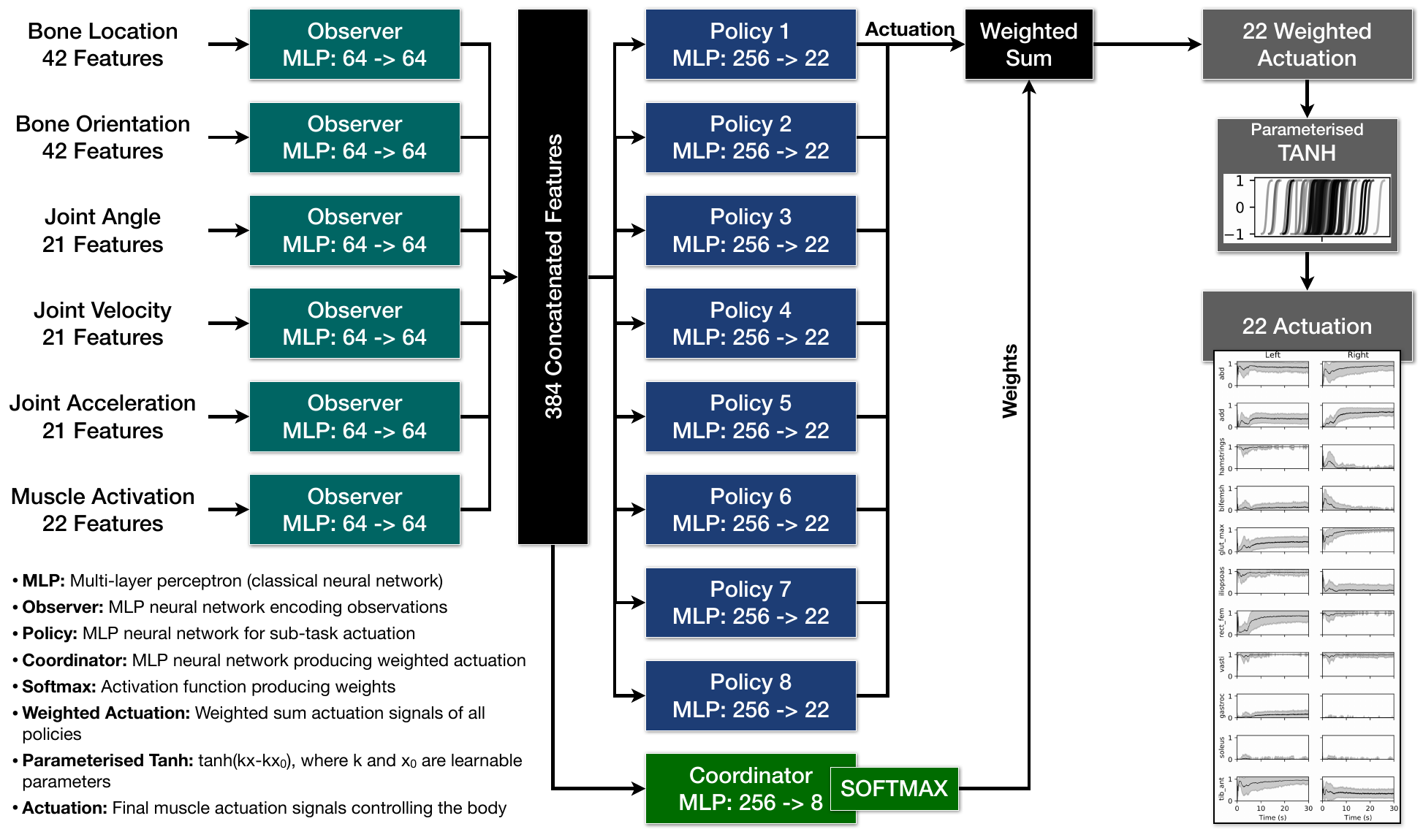}
    \caption{Neural network architecture of the actor in the trained agent. The actor architecture is adopted from~\cite{HossnyIskander2020_dontfall} while the action mapping using parameterised \texttt{tanh} is adopted from~\cite{HossnyEtal2020_PTANH}. The entire actor is trained first for 3000 episodes while only the coordinator sub-network is fine tuned later for 3000 more episodes.}
    \label{fig:actor}
\end{figure*}

\subsection{Coordinated Multi-Policy AI Agent}
We adopted the Deep Deterministic policy Gradient (DDPG) reinforcement learning~\cite{lillicrap2015continuous} method for training the AI agent controlling the digital musculoskeletal model. The DDPG AI agent is composed of two main multi-layer perceptron (MLP) neural networks; i.e. an actor and a critic. The actor serving as the controller which derives muscle excitation signals based on the observed input from the environment. The critic serves as an estimator to assess the efficacy of the actions produced by the actor given the feedback from the environment in terms of a single reward value. \textbf{The critic} neural network is a standard MLP with one input layer accepting 191 of concatenated inputs from the observation and action spaces, followed by three hidden layers with 256 neurons, and produces one output value (i.e. estimated episode reward). As shown in Figure~\ref{fig:actor}, \textbf{the actor} neural network, inspired by~\cite{HossnyIskander2020_dontfall}, featured six observation mapping sub-networks, eight policy sub-networks, one policy coordination sub-network and a parameterised \texttt{Tanh} layer~\cite{HossnyEtal2020_PTANH} for mapping produced actions to the action space. This architecture facilitates high flexibility in terms of enabling and disabling parts of the actor neural network during training. 

\subsubsection{Multiple observers}
The multi-head observer sub-networks enables controlling the gradient flow with updates suitable for the dynamic range of the input fed to each observer~\cite{HossnyIskander2020_dontfall}. Each observer consists of an input layer, two hidden layers (64 neuron es each) and an an output layer representing the encoded observation inn 64 values. A total of six observer sub-network were constructed to accommodate the observation space featuring bone positions (42 inputs), bone orientations (42 inputs), joint angles (21 inputs), joint velocity (21 inputs), joint acceleration (21 inputs) as well as muscle activation from previous state (22 inputs). The final representation of the observation is then encoded as a concatenation of the outputs from the the six observer sub-networks to produce a hidden feature vector of 384 values which will be fed to the subsequent policy and coordinator sub-networks. A \texttt{Tanh} activation function was used after each layer.

\subsubsection{Multiple policies}
The multi-policy sub-network allows the agent to learn different one skill per network using the proposed iterative training method. Each policy sub-network takes 384 features as a concatenated input from the aforementioned observer sub-networks. For each policy sub-network, the 348 input is fed into one hidden layer (256 neurons with \texttt{Tanh} activation). Each policy sub-network then produces a 22 excitation signal representing the action each policy would take given the observed state. The output layer used a linear activation function.

\subsubsection{Coordinator policy}
The role of the coordinator policy is to derive the action as a linear combination of the actions produced by the multiple policy sub-networks. It takes the concatenated 348 features produced by the observer sub-networks and feeds them into one hidden layer (256 neurons with \texttt{Tanh} activation). The coordination network produces a weighting factor for each action produced from the policy sub-networks via a 6-neurons \texttt{Softmax} layer. The six actions produced from the policy sub-networks are multiplied by the weighting produced from the \texttt{Softmax layer} to produce an approximated action. The approximated is then passed through a 22-neurons parameterised \texttt{Tanh} layer~\cite{HossnyEtal2020_PTANH} to produce the final action.

\subsection{Proposed DDPG training procedure}
The proposed training took place in two stages (3000 episodes each); i.e. iterative training and coordination refinement. The iterative training is responsible for training the observers, the coordination sub-network and one policy sub-network independently while locking the other policy sub-networks for 5000 simulation steps. Then the active policy sub-network is locked and the next policy sub-network is allowed to train. In doing so, the other policy sub-network serves as a regularising component and both the coordination and observer sub-networks learn to accommodate all policy sub-networks. 

\smallbreak
In the second training stage, all sub-networks in the actor were locked and only the coordination sub-network was allowed to train. This declares the learned policies as part of the environment and forces the coordination sub-network to learn the coordination patterns given the observed environment. 

\begin{figure*}
    \centering
    \subfigure[]{
        \includegraphics[height=.23\textheight]{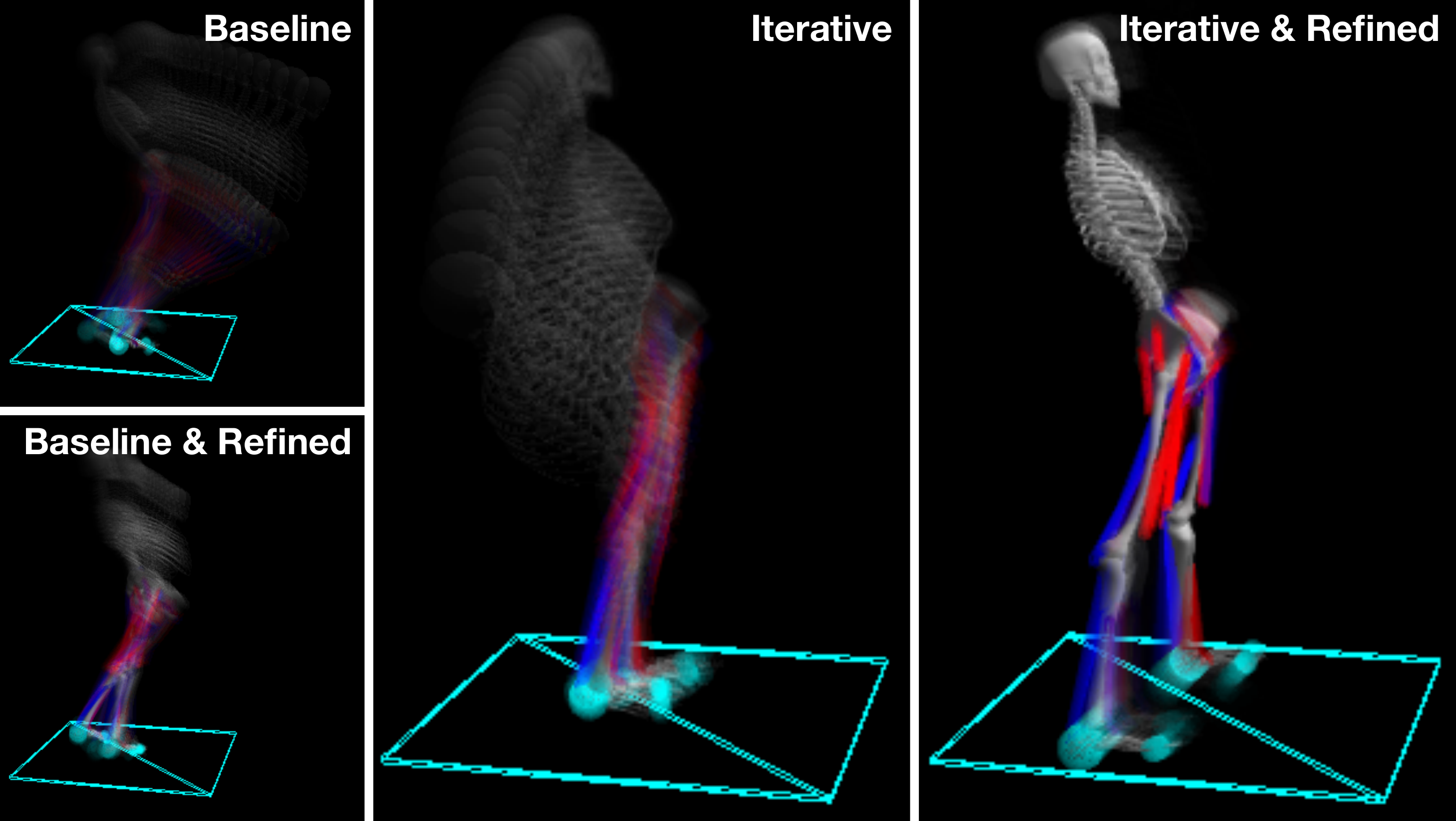}}
    \subfigure[]{
        \includegraphics[height=.23\textheight]{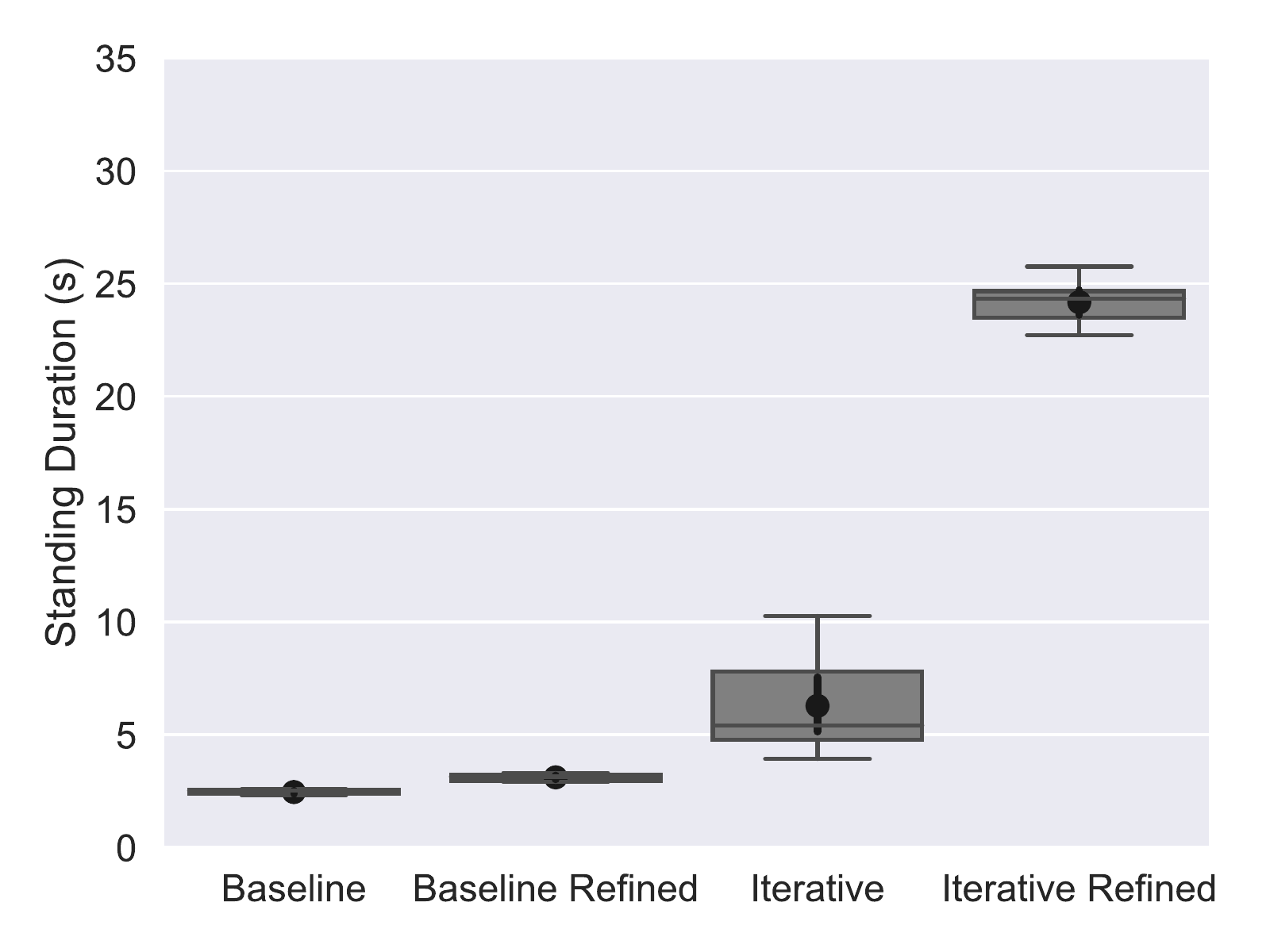}}

    \caption{Left: An overlaid time-lapsed visual representation of the AI agent's behaviour to compare between the brute force back-propagation (baseline) and the iterative training method (proposed). Muscle activation is colour coded (blue=0.0, purple=0.5 and red=1.0). Frames were loosely sampled for the \texttt{Iterative \& Refined} model to demonstrate the motion. Best viewed in colour. Right: A performance comparison of the same agent trained using the traditional and the proposed scheduled training. The comparison performance metric is the standing duration in seconds.}
    \label{fig:res}
\end{figure*}

\smallbreak
The training of the neural networks was optimised on 64-samples mini-batches using the \texttt{Adam} optimiser~\cite{kingma2014adam} with a learning rate, decay rate and first and second moments set to $1^{-4}$, $0.0$, $0.9, 0.999$, respectively. The DDPG update took place using soft update with update rate $\tau=5^{-3}$ and discounted reward $\gamma=0.99$. The experience replay buffer was allocated to accommodate $10^6$ experience samples.

\subsection{Baseline DDPG training procedure}
The baseline training took place in two stages. In the first stage, we trained the agent with all the sub-policy neural networks and coordinator at once for 3000 episodes. This is the typical training approach used in deep reinforcement learning. The second stage branched into two approaches (3000 episodes each); that is continuing the training as the first step or fine tuning the coordinator neural network only. This produced the agents we called \texttt{Baseline} and \texttt{Baseline Refined}, respectively. During fine tuning of the coordinator network, all other sub-networks were locked.

\subsection{Checkpoint selection}
For model selection, we adopted a strict selection criterion for the first stage by saving the model recording the highest instantaneous reward, highest average reward over the past 10 episodes and highest median reward over the past 10 episodes. This ensured selecting the model with highest sustainable reward. The nature of the problem dictated using such a strict criterion because during a fall (early termination), the actor's failed attempts to prevent falling results in altering the state of the actor neural network. This means, that even when a certain checkpoint does produce a good reward, it is not guaranteed that it will be saved by the time the episode ends. During the second stage of the training, i.e. coordination refinement, we relaxed the strict criteria to choose the best model based on the highest reward only. The reason behind this is that the baseline method failed to enhance the learned skills after the first few episodes which relying on not yet refined neural networks pre-trained from the first stage.

\subsection{Reproduciblity} For training, we chose an arbitrary seed of $107$ and tested over five different seeds in \{10638, 12190, 12944, 13734, 14798\}. The test seeds featured a balanced representation of 1's and 0's in the binary representation of each seed number $56\% \pm 12\%$. More technical details and the source code are available in the supplementary material.

\section{Results}
\label{sc:results}
The AI agent was trained on a single scenario which features the mannequin dropping from a $2 \pm 2$ centimetres onto a levelled ground platform. During testing we expanded the drop to range from 94 to 100 centimetres. We also tested the trained agent on inclined ground platform. All tests were conducted with five different seed values to ensure a wide variety of initial conditions. A visual comparison between the brute force backpropagation (baseline) and the proposed iterative training method is demonstrated in Figure~\ref{fig:res}-a.

\subsection{Coordinated posture stabilisation}
Iterative training of the AI agent did allow for more flexibility in terms of learning to achieve sub-goals (e.g. abduction and adduction of legs) in order to achieve the final goal of maintaining a standing posture as shown in Figure~\ref{fig:res}-b. The brute force back propagation baseline (\texttt{Baseline}) was not able to handle a standing posture for a duration more than 2 seconds after 3000 episodes. Extending the training of the \texttt{Baseline} to refine the coordination network while locking sub-actor networks did achieve an extra second of standing duration after additional 3000 episodes. The iterative learning (\texttt{Iterative}), on the other hand, did achieve a 50\% improvement over the \texttt{Refined Baseline} model with an average of 6 seconds. We then extended the training of the \texttt{Iterative} model to fine tune the coordination neural network which achieved a significant improvement of 300\% over the \texttt{Iterative} model.

\smallbreak


\subsection{Generalisation over perceptive perturbation noise}
\begin{figure}
    \centering
    \includegraphics[width=\linewidth]{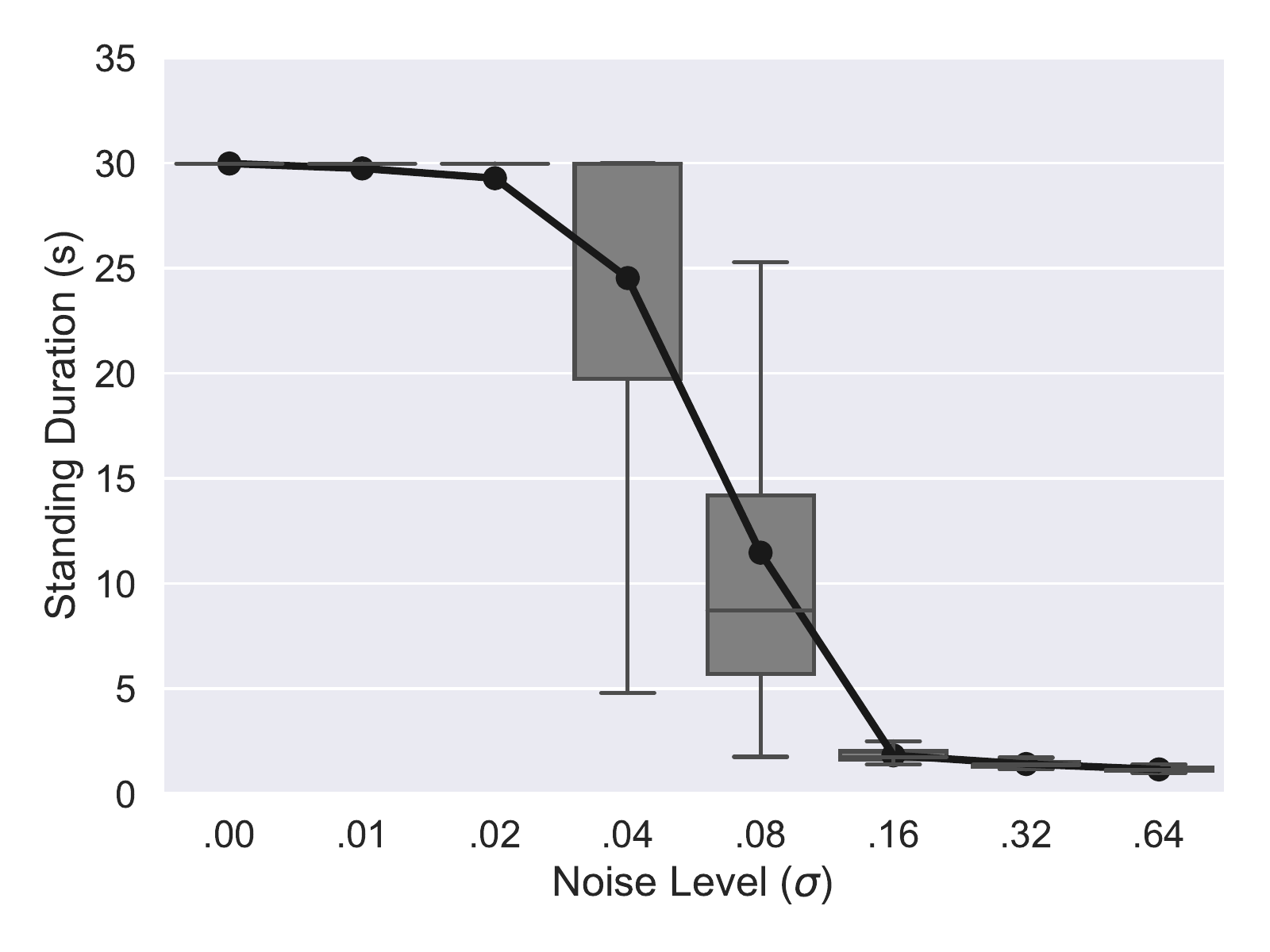}

    \caption{The effect of different state noise levels (perceived by the trained agent) on the posture stabilisation procedure.}
    \label{fig:res_noise}
\end{figure}

We also assessed the robustness of the trained agent against misleading observations. In order to achieve this, we added Gaussian noise $N(0, \sigma)$ with different levels, measured by the standard deviation $\sigma$, to the observation values obtained from the environment before engaging the control agent. We tested this on the best model \texttt{Refined Iterative} dropped from 6 cm height on a levelled platform. We increased the noise level exponentially with $\sigma=0.0$ (no noise) to $\sigma=0.64$. The physical interpretation of the added noise for each observation reading differs depending on the measured physical phenomenon. For angular measurements, velocities and accelerations, the noise levels ranges from $[-36.7, 36.7] \deg$. For spatial localisation of the rigid bodies, the noise levels range from $[-64, 64]$ centimetres. In terms of muscle activation, the added noise also translates to $\pm 0.64$ which is clipped to a minimum activation of 0.0 (no muscle activation) and 1.0 (fully activated muscle). The results shown in Figure~\ref{fig:res_noise} demonstrate that the trained agent is stable with $\pm 1.15 \deg$ angular noise, $\pm 2.0~\text{cm}$ spatial noise. The robustness drops logarithmically with $\{\pm 2.29, \pm 4.58\}\deg$ angular noise, $\{\pm 4.0, \pm 8.0 \}\text{cm}$ spatial noise. In this case, the agent can sustain a stable standing posture for 25 and 10 seconds, respectively. The robustness drops exponentially as the angular and spatial noise level increase to $\{\pm 9.17, \pm 36.67\} \deg$ and $\{\pm 8.0, \pm 64.0\} \text{cm}$, respectively (Supplementary Video 2).

\subsection{Generalisation over falling distance}
\begin{figure*}
    \centering
    \subfigure[Drop]{
    \includegraphics[width=1\linewidth]{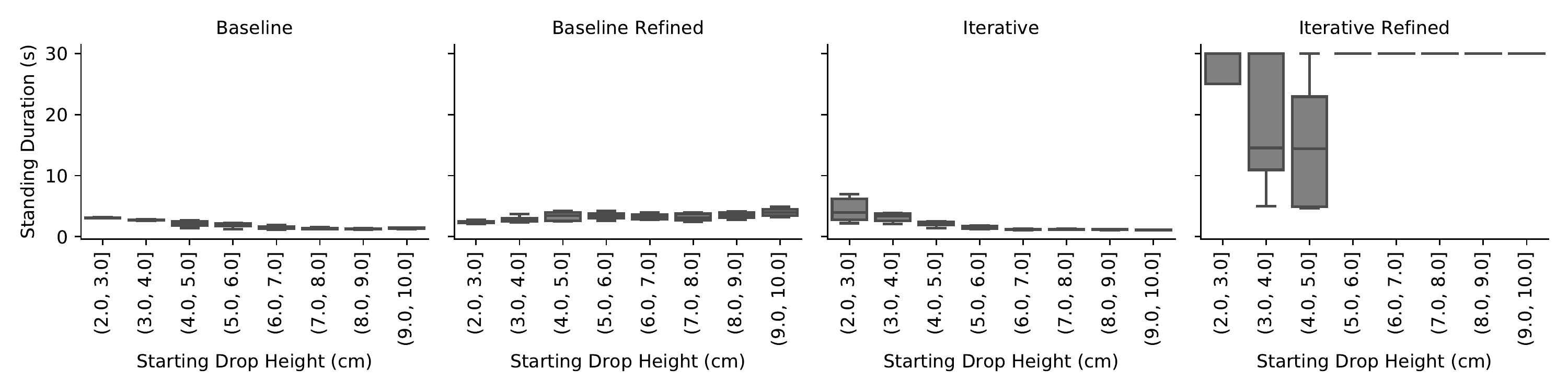}}\\
    \subfigure[Forward Inclination]{
    \includegraphics[width=1\linewidth]{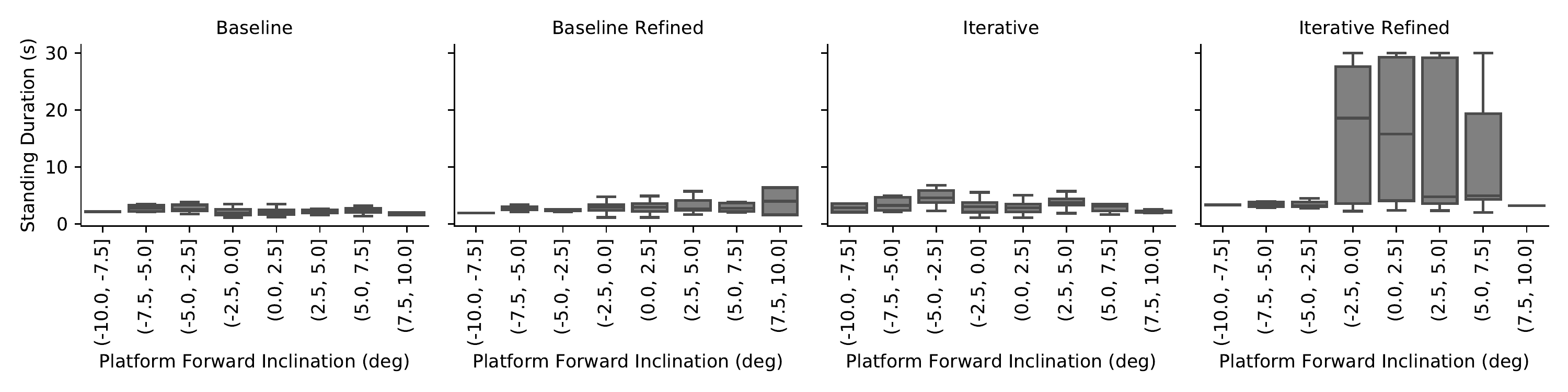}}\\
    \subfigure[Lateral Inclination]{\includegraphics[width=1\linewidth]{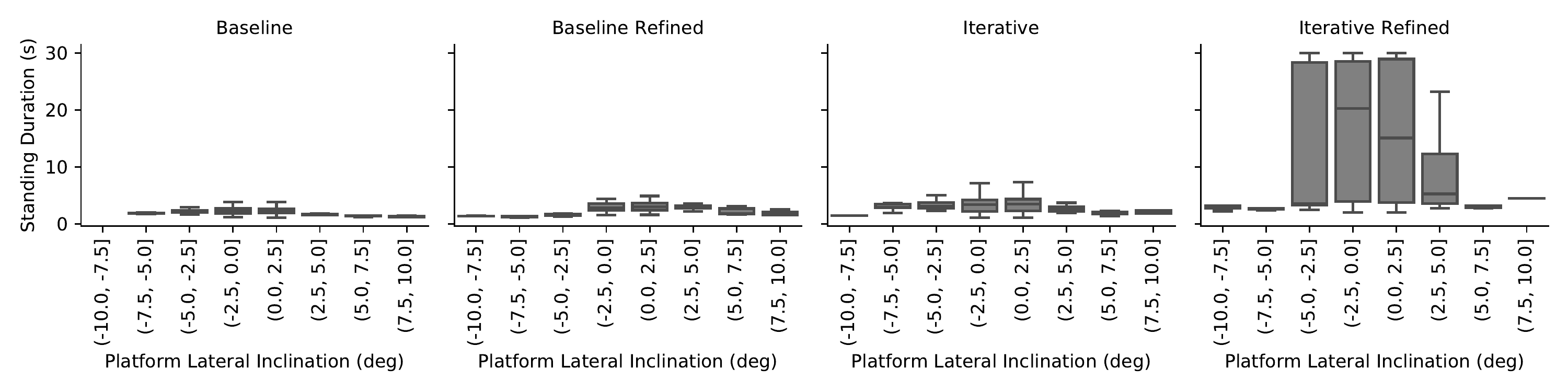}}

    \caption{The effect of different drop heights and forward/lateral ground inclination on the posture stabilisation procedure.}
    \label{fig:res_gen}
\end{figure*}

In order to assess the performance of the trained AI agent, we conducted testing with randomly generated drop distances while fixing the ground forward and lateral inclination to $0\deg$. As shown in Figure~\ref{fig:res_gen}-a, the iterative agent does outperform the baseline agents within a drop range of $[2, 4]$ but performs similarly as the drop distance increases. It is worth noting that fine tuning the coordination neural network does allow for better generalisation over drop distance as shown by comparing the box plots of \texttt{Baseline} and \texttt{Baseline Refined}. A significantly higher improvement can be observed by comparing the box plots of \texttt{Baseline} and \texttt{Baseline Refined} which is due to the iterative training of the sub-actors in the first 3000 episodes. The increased duration or stable posture for drop heights in $[6, 10]$ centimetres is attributed to the fact that the AI agent has more time to alter the landing posture by spreading the legs. This comes at the price of attempting to balance the body after bouncing off the ground. Although a similar pattern can be witnessed with the \texttt{Baseline Refined} agent, it could not maintain balance for more than 5 seconds. This was not achievable by the non-refined AI agents. This, in return, suggests that the \texttt{Iterative Refined} agent is capable of handling unforeseen scenarios. This is the main advantage of isolating sub-policies and shift the decision making to the coordination network. Videos of the generalisation behaviour are available in the supplementary material (Supplementary Video 3).

\subsection{Generalisation over inclined platforms}
The success in handling out-of-range drop heights prompted us to test the trained AI agent on different forward and lateral inclination platform setups. We tested the best model from each method in a setup where the musculoskeletal model drops from [0, 4] centimetres to land on an inclined ground platform. The forward/lateral inclination angle was randomly generated according to a normal distribution with a mean in $\{-2.5, 0.0, 2.5\} \deg$ with a standard deviation of $2.5 \deg$ to generate random inclination angle $\theta \in (-10, 10) \deg$. The drop height was adjusted accordingly with an offset of $\sin(\theta)/2$. The results shown in Figure~\ref{fig:res_gen}-b, Figure~\ref{fig:res_gen}-c demonstrate the superiority of iterative training with coordination fine tuning especially when $\theta \in (-5.0, 5.0] \deg$. This demonstrate the generalisation capability of the trained AI agent and the effect of refining the coordination policy (Supplementary Video 4). Furthermore, the trained agent also managed to generalise on obliquely inclined platforms where lateral and forward inclination angle $\theta \in [-2.5, 2.5]$ (Supplementary Video 5). 


\smallbreak
% ...

\begin{figure*}
    \centering
    \subfigure[Policy 1: CW Rotation]{
        \includegraphics[trim=10cm 6.5cm 10.5cm 3.25cm, clip, width=.23\linewidth]{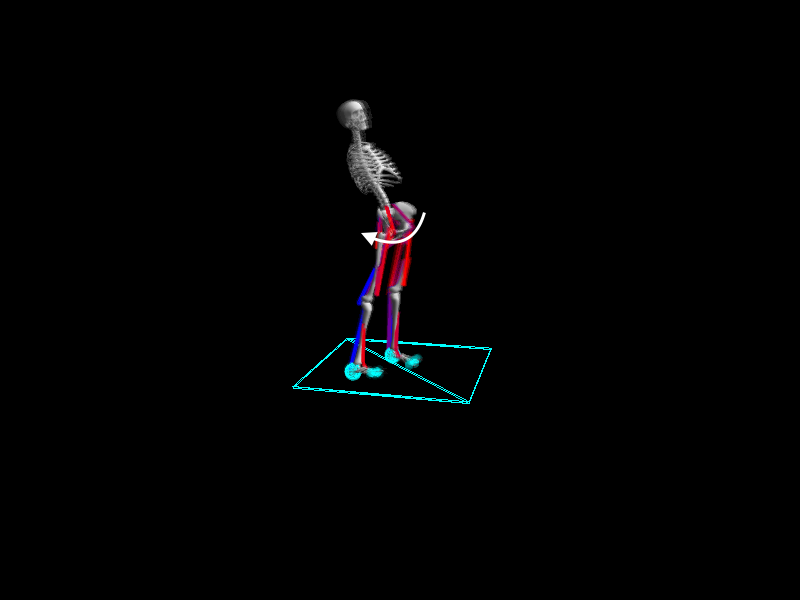}}
    \subfigure[Policy 2: Lean Backward]{
        \includegraphics[trim=10cm 6.5cm 10.5cm 3.25cm, clip, width=.23\linewidth]{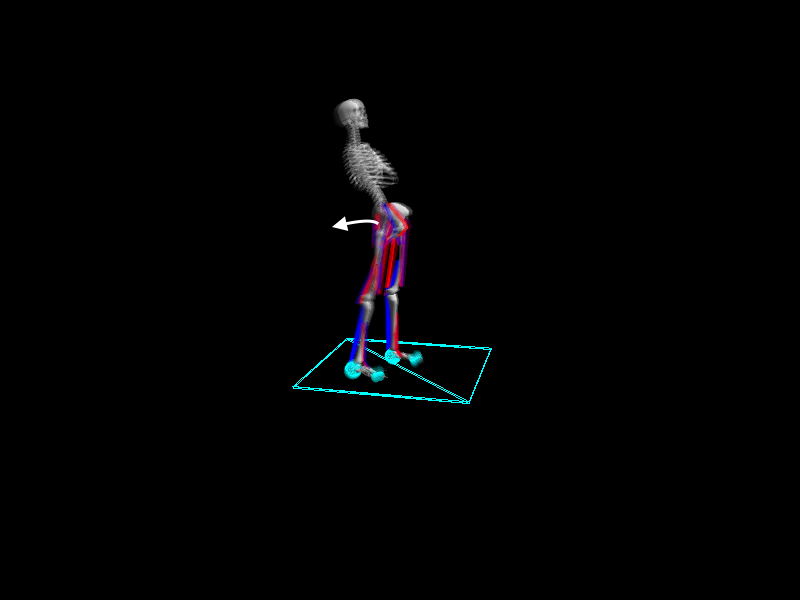}}
    \subfigure[Policy 3: CCW Rotation]{
        \includegraphics[trim=10cm 6.5cm 10.5cm 3.25cm, clip, width=.23\linewidth]{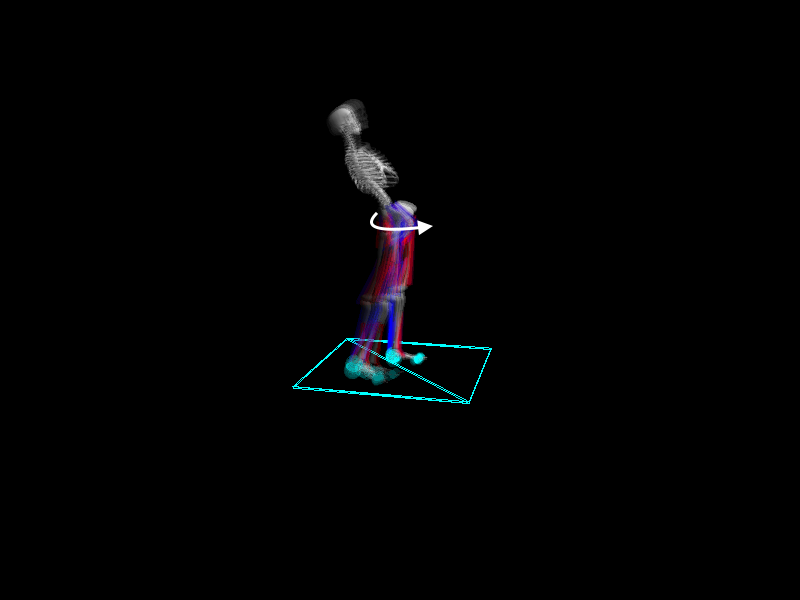}}
    \subfigure[Policy 4: Left Abduction]{
        \includegraphics[trim=10cm 6.5cm 10.5cm 3.25cm, clip, width=.23\linewidth]{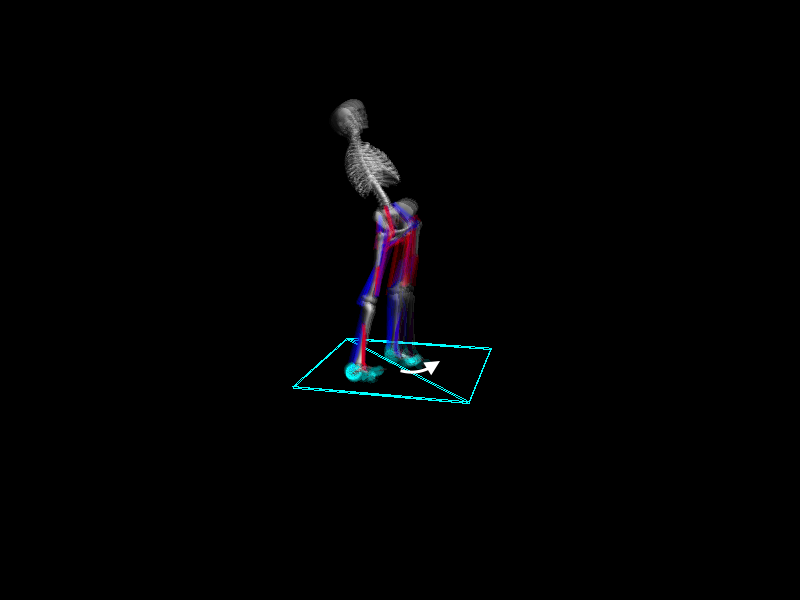}}\\
    \subfigure[Policy 5: Left Adduction]{
        \includegraphics[trim=10cm 6.5cm 10.5cm 3.25cm, clip, width=.23\linewidth]{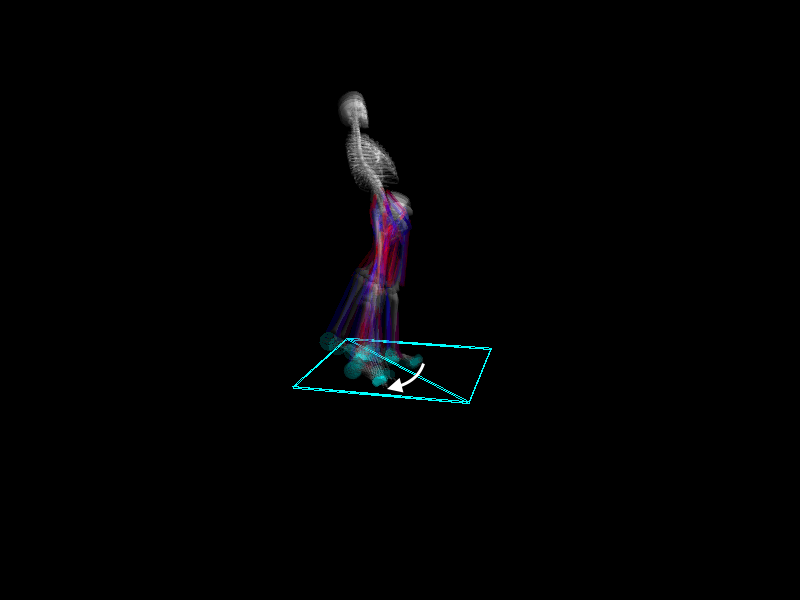}}
    \subfigure[Policy 6: Right Adduction]{
        \includegraphics[trim=10cm 6.5cm 10.5cm 3.25cm, clip, width=.23\linewidth]{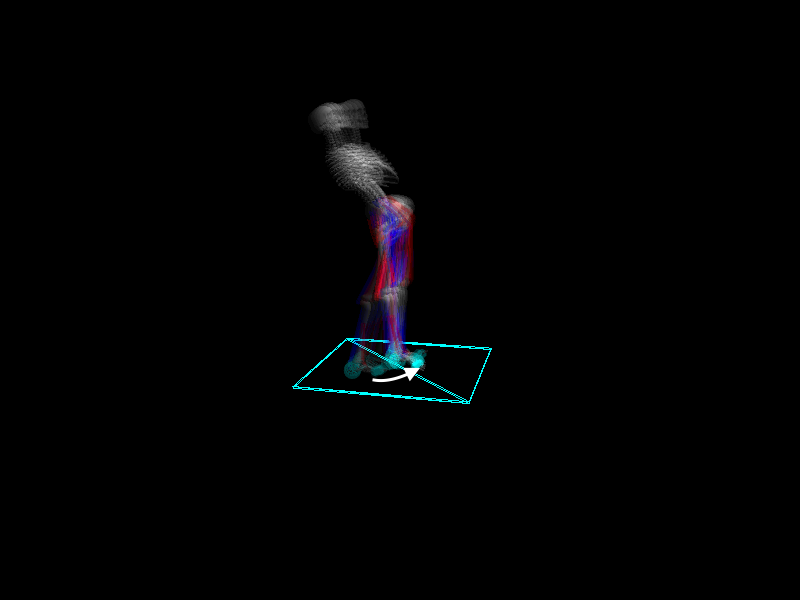}}
    \subfigure[Policy 7: Thrust Forward]{
        \includegraphics[trim=10cm 6.5cm 10.5cm 3.25cm, clip, width=.23\linewidth]{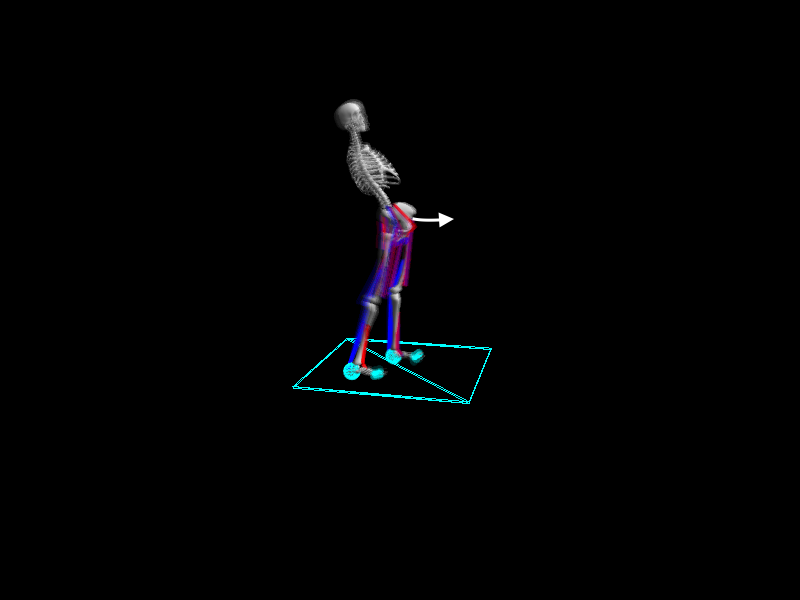}}
    \subfigure[Policy 8: Right Abduction]{
        \includegraphics[trim=10cm 6.5cm 10.5cm 3.25cm, clip, width=.23\linewidth]{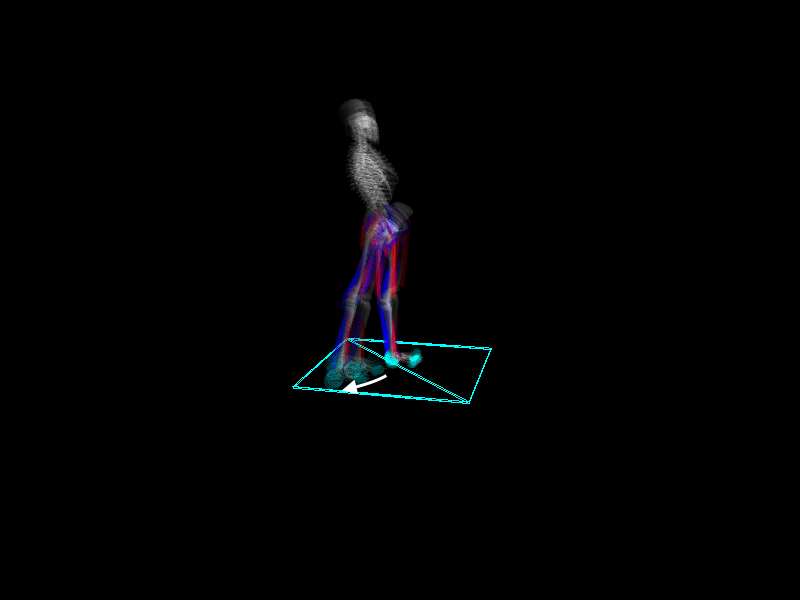}}
    \caption{An  overlaid  visual  representation  of  the learnt skills required to maintain a standing posture. Muscle activation is colour coded (blue=0.0, purple=0.5and red=1.0). Best viewed in colour.}
    \label{fig:policies}
\end{figure*}

\begin{figure*}
    \centering
    \subfigure[Iterative Only]{
        \includegraphics[width=.45\linewidth]{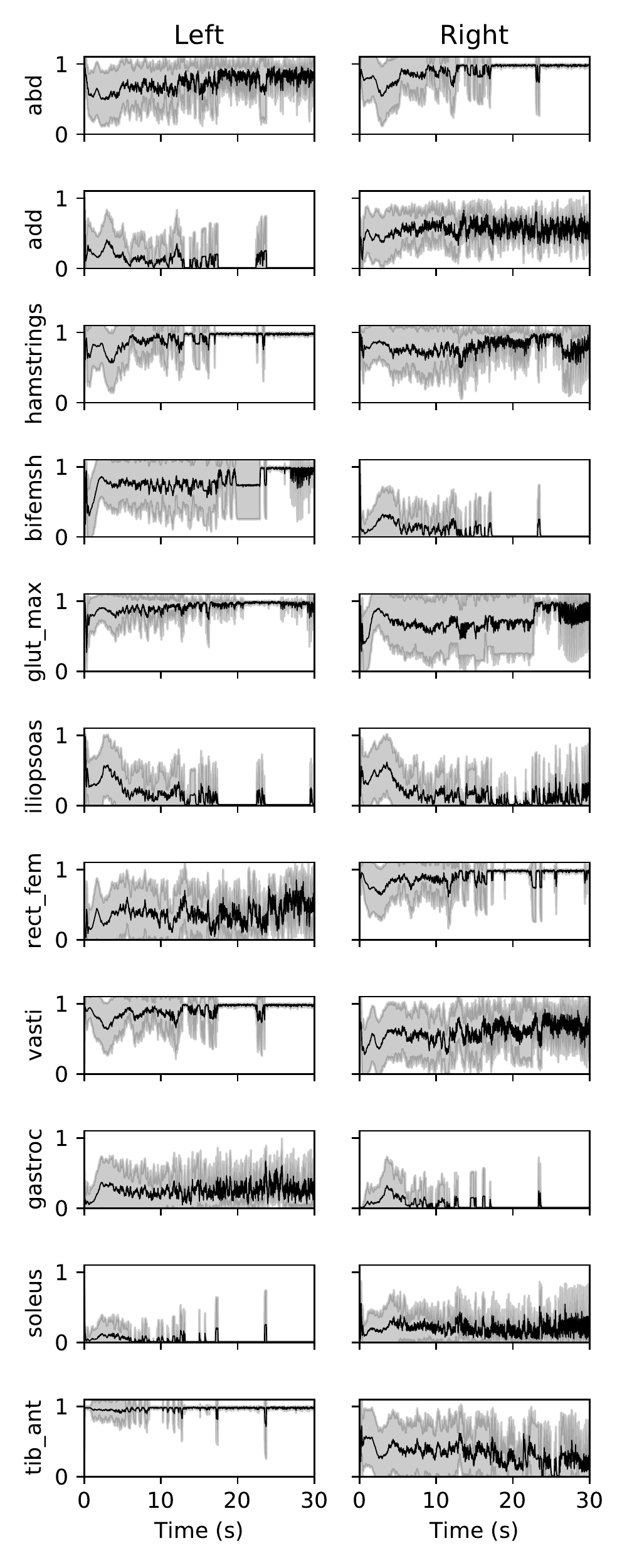}}
    \subfigure[Iterative With Refined Coordination]{
        \includegraphics[width=.45\linewidth]{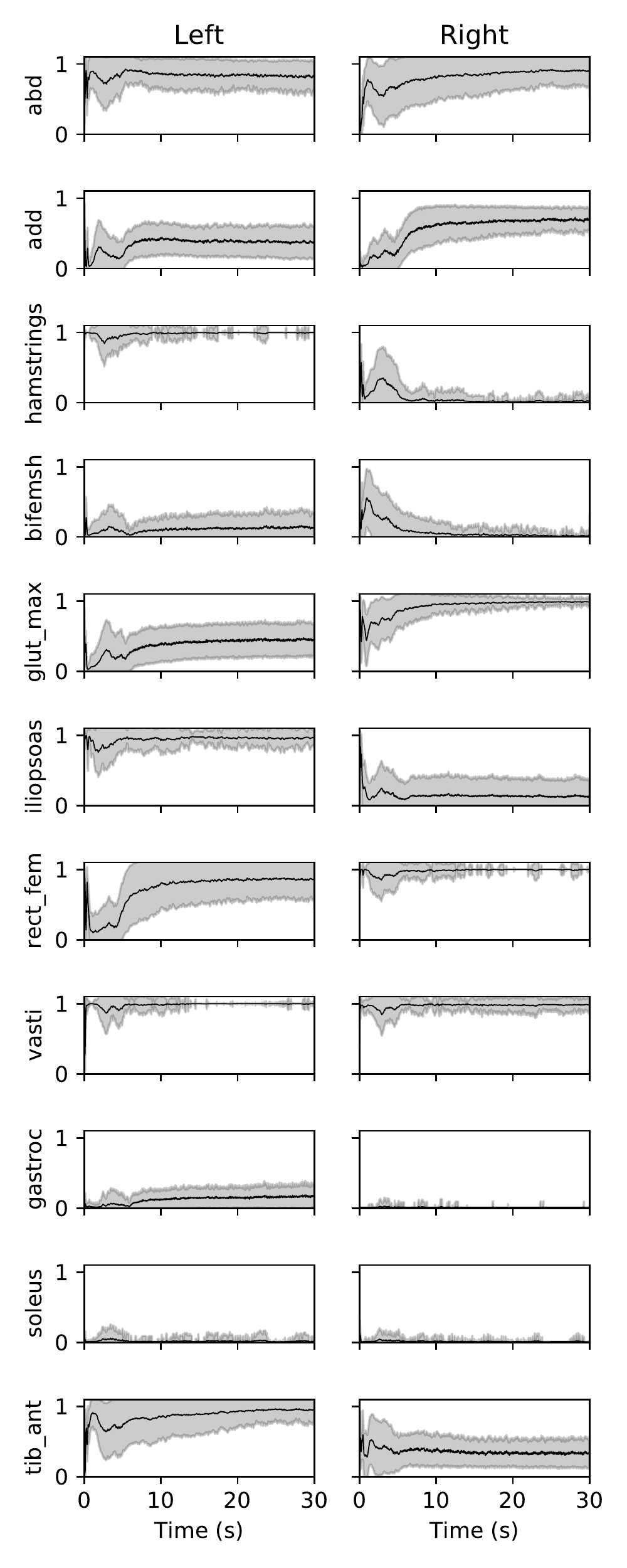}}
    
    \caption{Comparison between muscle activation signals of the proposed method with and without coordination refinement.}
    \label{fig:res_muscact}
\end{figure*}

\section{Discussion}
\label{sc:disc}
\smallbreak\smallbreak
During the design stage of this study, we hypothesised that the AI agent will be able to subdivide the task of maintaining a standing posture into 8 different sub-tasks spanning four directions which each leg can move along (i.e. forward, backward, left and right). We also hypothesised that each of these skills cannot solely maintain a standing posture. Yet the coordination between these skills should allow the AI agent to excite the lower body muscles in order to maintain balance. 

\smallbreak
Upon closer inspection of the learnt sub-tasks, the AI agent did indeed learn to identify and solve several sub-tasks. However, the learnt policies do not exhibit the anticipated sub-tasks we hypothesised. As shown in Figure~\ref{fig:policies}, the learnt skills fall into three categories; that is leg abduction, leg adduction and weight shift. The weight shifting sub-policies are also split into pelvis rotation (Policies 1 and 3) and leaning backward and thrusting forward (Policies 2 and 7). Both rotation directions used the left leg as the pivot point while adjusting the right leg to perform the rotation. Both left and right abduction policies were mainly engaged to widen the standing base (Policies 4 and 8). The adduction policies were the most unstable and yet crucial for performing a conservative abduction (Policies 5 and 6). The rotation policies were engaged in cooperation with right leg adduction (Policy 6) to shift the weight to the left and right in order to maintain lateral balance. Videos of learnt skills are available in the supplementary material (Supplementary Video 6).

\smallbreak
It is worth noting that while symmetry was observed in the policies dedicated to left and right leg, some policies appear to have matured for one leg more than their counterparts for the other leg. For instance, while Policy 8 demonstrate a successful lateral step using right leg abduction, Policy 4 does include a CW rotation. Another example is the difference in policies used during maintaining anterior/posterior balance where the agent thrusts forward (Policy 7) instead of leaning (like in Policy 2). This raises two important points. \textbf{Firstly}, the coordination refinement is an extremely important step during training. This has already been demonstrated by the results illustrated in Figure~\ref{fig:res}-b. This is also demonstrated in Figure~\ref{fig:res_muscact} which shows the effect of fine tuning the coordination policy on the stability of the assumed posture. This is because without fine tuning the coordination policy changes along with all the other policies; i.e. observers, policies and action mapping policies. Fine tuning the coordination policy alone allows the AI agent to focus only on deriving the linear combination of the actuation signals produced by the policies. \textbf{The second point} is related to the apparent evidence that the number of individual policies is not enough to exhibit all degrees of freedom the AI agent can exploit in order to achieve a balanced standing posture. This is clearly demonstrated in Policy 3 where the agent takes a small forward step with the right leg during a CCW pelvis rotation. 

\smallbreak
The coordination policy does serve a long term goal beyond maintaining a prolonged standing posture; i.e. explanation of produced actions. The use of a coordination policy allowed the AI agent to develop an internal finite state machine of the learned policies. While this allows choosing between policies, in practice it derives a linear combination of the actuation signals produced by the learnt policies. We investigated the possibility of training the coordination policy to produce discrete choices but the lack of backward derivatives for the \texttt{argmax} operator rendered this option infeasible. Another point to note is that the linear combination derived by the coordination policy assumes each actuation signal from each policy as a single vector and thus all actuation signals are scaled using the same weighting factor. This opens an important room for improvement if the coordination network is allowed to derive the linear combination of each muscle independently. However, this may affect the achieved visibility of the policy engagement procedure.

\section{Conclusions}
\label{sc:conc}
Allowing the AI agent to learn each policy as well as the coordination independently serves two important roles. First, it enables the AI to develop the finite state model for solving the problem. Second, it provides visibility over the decision making process the AI agent does when choosing one policy over another. The trained finite state model is an important step in expanding this research to assess the ergonomics of assembly tasks. While portable ergonomic assessment tools (e.g. \citep{abobakr2019rgb}) do provide a convenient assessment, thee real challenge lies in assessing this before hand. Combined with the principle of energy conservation (i.e. minimising effort), a trained RL agent would provide an insight on how an operator is likely to behave in different setups and thus conducting the ergonomic assessment before deployment. This can also expand to cover visual ergonomics using ocular biomechanics~\cite{iskander2020ocular} as well as the impact of the spatial location of visual stimuli on motivated rotations of the neck and the spine~\citep{iskander2017sim,iskander2018biomech} in real world as well as in virtual environments~\citep{iskander2018review,iskander2019using}. The combination of biomechanics and reward-driven AI can also provide a data generation pipeline for MOCAP data of animals which cannot accommodate markers placed on their bodies (e.g. cats) as well as animals in dangerous setups~\cite{saleh2018effective,haggag2015body}.

\smallbreak
It is important to note that a PD controller was not used at any stage of the training and testing. However, we would like to emphasise that complex control tasks such as maintaining biomechanic balance and locomotion of bipedal entities require several granularity levels of control. In theory, machine learning can provide a solution to several complex problems given enough compute power according to the universal approximation theorem~\cite{Csji2001ApproximationWA}. However, in practice narrowing the scope of the problem to fine tune parameters of existing models may prove more practical and cost effective. The methodology we presented in this work embraces this approach and articulates the advantage of solving multiple narrow-scoped tasks in contrast to solving the overall goal at once. Perhaps the recent results from Seoul National University on bipedal locomotion~\cite{MASS2019} serves as another testimony of this argument. In their work, Lee et al. introduced an interesting integration of classic control and deep reinforcement learning. Particularly, they used a PD controller to serve as a catalyst during the training of two deep neural networks which mimics a reference motion (i.e. input by the user) and the muscle coordination needed to produce this motion. This highlights an important point; that is as the field of artificial intelligence (AI) research progresses towards general intelligence (AGI), it is important to acknowledge the efficacy of training procedures and thus expand the discipline of machine learning to accommodate machine teaching as well.





%




% ...

% ...



\begin{thebibliography}{10}

\bibitem{Hidenori200618}
K.~Hidenori and Y.~Jiang, ``A pid model of human balance keeping,'' {\em IEEE
  Control Systems}, vol.~26, no.~6, pp.~18--23, 2006.
\newblock cited By 23.

\bibitem{novacheck1998biomechanics}
T.~F. Novacheck, ``The biomechanics of running,'' {\em Gait \& posture},
  vol.~7, no.~1, pp.~77--95, 1998.

\bibitem{bohannon1997comfortable}
R.~W. Bohannon, ``Comfortable and maximum walking speed of adults aged 20—79
  years: reference values and determinants,'' {\em Age and ageing}, vol.~26,
  no.~1, pp.~15--19, 1997.

\bibitem{serra2015postural}
P.~Serra-A{\~N}{\'o}, L.~L{\'o}pez-Bueno, X.~Garc{\'\i}a-Mass{\'o}, M.~T.
  Pellicer-Chenoll, and L.-M. Gonz{\'a}lez, ``Postural control mechanisms in
  healthy adults in sitting and standing positions,'' {\em Perceptual and motor
  skills}, vol.~121, no.~1, pp.~119--134, 2015.

\bibitem{park2016coordination}
E.~Park, H.~Reimann, and G.~Sch{\"o}ner, ``Coordination of muscle torques
  stabilizes upright standing posture: an ucm analysis,'' {\em Experimental
  brain research}, vol.~234, no.~6, pp.~1757--1767, 2016.

\bibitem{miller1999mechanical}
C.~A. Miller and M.~C. Verstraete, ``A mechanical energy analysis of gait
  initiation,'' {\em Gait \& posture}, vol.~9, no.~3, pp.~158--166, 1999.

\bibitem{barroso2017combining}
F.~O. Barroso, D.~Torricelli, F.~Molina-Rueda, I.~M. Alguacil-Diego,
  R.~Cano-de-la Cuerda, C.~Santos, J.~C. Moreno, J.~C. Miangolarra-Page, and
  J.~L. Pons, ``Combining muscle synergies and biomechanical analysis to assess
  gait in stroke patients,'' {\em Journal of Biomechanics}, vol.~63,
  pp.~98--103, 2017.

\bibitem{delp2007opensim}
S.~L. Delp, F.~C. Anderson, A.~S. Arnold, P.~Loan, A.~Habib, C.~T. John,
  E.~Guendelman, and D.~G. Thelen, ``Opensim: open-source software to create
  and analyze dynamic simulations of movement,'' {\em IEEE transactions on
  biomedical engineering}, vol.~54, no.~11, pp.~1940--1950, 2007.

\bibitem{anybody2000}
J.~Rasmussen, ``The anybody project - modeling human motion by computer,''
  2000.

\bibitem{MASS2019}
S.~Lee, K.~Lee, M.~Park, and J.~Lee, ``Scalable muscle-actuated human
  simulation and control,'' {\em ACM Transactions on Graphics}, vol.~37, 2019.

\bibitem{delp1990interactive}
S.~L. Delp, J.~P. Loan, M.~G. Hoy, F.~E. Zajac, E.~L. Topp, and J.~M. Rosen,
  ``An interactive graphics-based model of the lower extremity to study
  orthopaedic surgical procedures,'' {\em IEEE Transactions on Biomedical
  engineering}, vol.~37, no.~8, pp.~757--767, 1990.

\bibitem{delp1992force}
S.~L. Delp and F.~E. Zajac, ``Force-and moment-generating capacity of
  lower-extremity muscles before and after tendon lengthening.,'' {\em Clinical
  orthopaedics and related research}, vol.~284, pp.~247--259, 1992.

\bibitem{kidzinski2020artificial}
{\L}.~Kidzi{\'n}ski, C.~Ong, S.~P. Mohanty, J.~Hicks, S.~Carroll, B.~Zhou,
  H.~Zeng, F.~Wang, R.~Lian, H.~Tian, {\em et~al.}, ``Artificial intelligence
  for prosthetics: Challenge solutions,'' in {\em The NeurIPS'18 Competition},
  pp.~69--128, Springer, 2020.

\bibitem{iskander2018biomech}
J.~Iskander, M.~Hossny, and S.~Nahavandi, ``Biomechanical analysis of eye
  movement in virtual environments: A validation study,'' in {\em Systems, Man,
  and Cybernetics (SMC), 2018 IEEE International Conference on}, p.~Accepted,
  IEEE, 2018.

\bibitem{iskander2019using}
J.~Iskander, M.~Hossny, and S.~Nahavandi, ``{Using biomechanics to investigate
  the effect of VR on eye vergence system},'' {\em Applied Ergonomics},
  vol.~81, p.~102883, 2019.

\bibitem{iskander2018review}
J.~Iskander, M.~Hossny, and S.~Nahavandi, ``A review on ocular biomechanic
  models for assessing visual fatigue in virtual reality,'' {\em IEEE Access},
  vol.~6, pp.~19345--19361, 2018.

\bibitem{iskander2017sim}
J.~Iskander, M.~Hossny, and S.~Nahavandi, ``Simulating eye-head coordination
  during smooth pursuit using an ocular biomechanic model,'' in {\em Systems,
  Man, and Cybernetics (SMC), 2017 IEEE International Conference on},
  pp.~3356--3361, IEEE, 2017.

\bibitem{abobakr2017fall}
A.~Abobakr, M.~Hossny, and S.~Nahavandi, ``A skeleton-free fall detection
  system from depth images using random decision forest,'' {\em IEEE Systems
  Journal}, vol.~PP, no.~99, pp.~1--12, 2018.

\bibitem{abobakr2018lstmfall}
A.~{Abobakr}, M.~{Hossny}, H.~{Abdelkader}, and S.~{Nahavandi}, ``Rgb-d fall
  detection via deep residual convolutional lstm networks,'' pp.~1--7, 2018.

\bibitem{kidzinski2018learning}
{\L}.~Kidzi{\'n}ski, S.~P. Mohanty, C.~F. Ong, J.~L. Hicks, S.~F. Carroll,
  S.~Levine, M.~Salath{\'e}, and S.~L. Delp, ``Learning to run challenge:
  Synthesizing physiologically accurate motion using deep reinforcement
  learning,'' in {\em The NIPS'17 Competition: Building Intelligent Systems},
  pp.~101--120, Springer, 2018.

\bibitem{Geijtenbeek}
T.~Geijtenbeek, M.~van~de Panne, and A.~F. van~der Stappen, ``Flexible
  muscle-based locomotion for bipedal creatures,'' {\em ACM Trans. Graph.},
  vol.~32, Nov. 2013.

\bibitem{HossnyIskander2020_dontfall}
M.~Hossny and J.~Iskander, ``Just don't fall: An ai agent's learning journey
  towards posture stabilisation,'' {\em AI}, vol.~1, no.~2, pp.~286--298.,
  2020.

\bibitem{Bottaro2005588}
A.~Bottaro, M.~Casadio, P.~Morasso, and V.~Sanguineti, ``Body sway during quiet
  standing: Is it the residual chattering of an intermittent stabilization
  process?,'' {\em Human Movement Science}, vol.~24, no.~4, pp.~588--615, 2005.
\newblock cited By 135.

\bibitem{stanley2017neuroevolution}
K.~O. Stanley, ``Neuroevolution: A different kind of deep learning,'' {\em
  Obtenido de https://www. oreilly.
  com/ideas/neuroevolution-a-different-kindof-deep-learning el}, vol.~27,
  no.~04, p.~2019, 2017.

\bibitem{sutton2000policy}
R.~S. Sutton, D.~A. McAllester, S.~P. Singh, and Y.~Mansour, ``Policy gradient
  methods for reinforcement learning with function approximation,'' in {\em
  Advances in neural information processing systems}, pp.~1057--1063, 2000.

\bibitem{HANSEN2006CMAe}
N.~Hansen, ``The cma evolution strategy: a comparing review,'' {\em Towards a
  new evolutionary computation}, p.~75–102, 2006.

\bibitem{mnih2013playing}
V.~Mnih, K.~Kavukcuoglu, D.~Silver, A.~Graves, I.~Antonoglou, D.~Wierstra, and
  M.~Riedmiller, ``Playing atari with deep reinforcement learning,'' {\em arXiv
  preprint arXiv:1312.5602}, 2013.

\bibitem{lillicrap2015continuous}
T.~P. Lillicrap, J.~J. Hunt, A.~Pritzel, N.~Heess, T.~Erez, Y.~Tassa,
  D.~Silver, and D.~Wierstra, ``Continuous control with deep reinforcement
  learning,'' {\em arXiv preprint arXiv:1509.02971}, 2015.

\bibitem{silver2014deterministic}
D.~Silver, G.~Lever, N.~Heess, T.~Degris, D.~Wierstra, and M.~Riedmiller,
  ``Deterministic policy gradient algorithms,'' in {\em International
  Conference on Machine Learning}, pp.~387--395, 2014.

\bibitem{haarnoja2018soft}
T.~Haarnoja, A.~Zhou, P.~Abbeel, and S.~Levine, ``Soft actor-critic: Off-policy
  maximum entropy deep reinforcement learning with a stochastic actor,'' in
  {\em International Conference on Machine Learning}, pp.~1861--1870, 2018.

\bibitem{haarnoja2018softarx}
T.~Haarnoja, A.~Zhou, K.~Hartikainen, G.~Tucker, S.~Ha, J.~Tan, V.~Kumar,
  H.~Zhu, A.~Gupta, P.~Abbeel, {\em et~al.}, ``Soft actor-critic algorithms and
  applications,'' {\em arXiv preprint arXiv:1812.05905}, 2018.

\bibitem{PPO2017}
J.~Schulman, F.~Wolski, P.~Dhariwal, A.~Radford, and O.~Klimov, ``Proximal
  policy optimization algorithms,'' {\em arXiv:1707.06347}, 2017.

\bibitem{HossnyEtal2020_PTANH}
M.~Hossny, J.~Iskander, M.~Attia, and K.~Saleh, ``Refined continuous control of
  ddpg actors via parametrised activation,'' {\em arXiv:2006.02818 [cs.LG]},
  2020.

\bibitem{seth2018opensim}
A.~Seth, J.~L. Hicks, T.~K. Uchida, A.~Habib, C.~L. Dembia, J.~J. Dunne, C.~F.
  Ong, M.~S. DeMers, A.~Rajagopal, M.~Millard, {\em et~al.}, ``Opensim:
  Simulating musculoskeletal dynamics and neuromuscular control to study human
  and animal movement,'' {\em PLoS computational biology}, vol.~14, no.~7,
  p.~e1006223, 2018.

\bibitem{brockman2016openai}
G.~Brockman, V.~Cheung, L.~Pettersson, J.~Schneider, J.~Schulman, J.~Tang, and
  W.~Zaremba, ``Openai gym,'' {\em arXiv preprint arXiv:1606.01540}, 2016.

\bibitem{ong2017predictive}
C.~F. Ong, T.~Geijtenbeek, J.~L. Hicks, and S.~L. Delp, ``Predictive
  simulations of human walking produce realistic cost of transport at a range
  of speeds,'' in {\em Proceedings of the 16th International Symposium on
  Computer Simulation in Biomechanics}, pp.~19--20, 2017.

\bibitem{arnold2010model}
E.~M. Arnold, S.~R. Ward, R.~L. Lieber, and S.~L. Delp, ``A model of the lower
  limb for analysis of human movement,'' {\em Annals of biomedical
  engineering}, vol.~38, no.~2, pp.~269--279, 2010.

\bibitem{thelen2003adjustment}
D.~G. Thelen, ``Adjustment of muscle mechanics model parameters to simulate
  dynamic contractions in older adults,'' {\em Journal of biomechanical
  engineering}, vol.~125, no.~1, pp.~70--77, 2003.

\bibitem{hunt1975coefficient}
K.~Hunt and F.~Crossley, ``Coefficient of restitution interpreted as damping in
  vibroimpact,'' {\em Journal of Applied Mechanics}, vol.~42, no.~2,
  pp.~440--445, 1975.

\bibitem{kingma2014adam}
D.~P. Kingma and J.~Ba, ``Adam: A method for stochastic optimization,'' {\em
  arXiv:1412.6980}, 2014.

\bibitem{abobakr2019rgb}
A.~Abobakr, D.~Nahavandi, M.~Hossny, J.~Iskander, M.~Attia, S.~Nahavandi, and
  M.~Smets, ``Rgb-d ergonomic assessment system of adopted working postures,''
  {\em Applied Ergonomics}, vol.~80, pp.~75--88, 2019.

\bibitem{iskander2020ocular}
J.~Iskander and M.~Hossny, ``An ocular biomechanics environment for
  reinforcement learning,'' {\em arXiv:2008.05088}, 2020.

\bibitem{saleh2018effective}
K.~Saleh, M.~Hossny, and S.~Nahavandi, ``Effective vehicle-based kangaroo
  detection for collision warning systems using region-based convolutional
  networks,'' {\em Sensors}, vol.~18, no.~6, p.~1913, 2018.

\bibitem{haggag2015body}
H.~Haggag, M.~Hossny, S.~Nahavandi, S.~Haggag, and D.~Creighton, ``Body parts
  segmentation with attached props using rgb-d imaging,'' in {\em Digital Image
  Computing: Techniques and Applications (DICTA), 2015 International Conference
  on}, pp.~1--8, IEEE, 2015.

\bibitem{Csji2001ApproximationWA}
B.~C. Cs{\'a}ji, ``Approximation with artificial neural networks,'' 2001.

\end{thebibliography}

\end{document}